# Exploring persuasive Interactions with generative social Robots – An experimental framework

**Stephan Vonschallen**

ZHAW Zurich University of Applied Sciences, FHNW University of Applied Sciences and Arts Northwestern Switzerland, & Bielefeld University

Corresponding author email: stephan.vonschallen@zhaw.ch

**Larissa Julia Corina Finsler**

FHNW University of Applied Sciences and Arts Northwestern Switzerland

**Theresa Schmiedel**

ZHAW Zurich University of Applied Sciences

**Friederike Eyssel**

Bielefeld University

**ABSTRACT**

Integrating generative AI like Large Language Model into social robots has significantly enhanced their ability to engage in natural, human-like communication. This study introduces a methodological approach to investigate the persuasive capabilities of these generative social robots. We developed an experimental framework focused on decision-making and tested its feasibility through a real-world pilot experiment that varied robot appearance and self-knowledge. Using a qualitative approach, we analyzed interaction quality, persuasion effectiveness, and the robot's communicative strategies. Participants generally experienced the interaction positively, describing the robot as competent, friendly, and supportive, though they noted technical limitations such as delayed responses and occasional speech recognition errors. Persuasiveness was found to be highly context-dependent and shaped by the robot's behavior: participants responded well to polite, reasoned suggestions and expressive gestures, but emphasized the need for more personalized, context-aware arguments and clearer social roles. These findings suggest that while GSRs can influence user decisions, their persuasion effectiveness depends on communicative nuance and contextual relevance. We propose refinements to the framework to further explore persuasive dynamics between GSRs and human users.

# 1 INTRODUCTION

Given rapid developments in Large Language Models (LLMs), so called generative social robots (GSRs) [28] that utilize generative AI to express verbal and non-verbal behavior can now communicate in a highly natural manner [5]. While this development offers new opportunities for social robots to engage with users, it also comes with risks. Among these are a potential loss of user autonomy, the danger of being discriminated against, or being manipulated by technology into changing one's attitudes or behavior [33].

To prevent undesired outcomes like manipulation or deception, it is key to understand how GSRs can shape user attitudes and behaviors through communication, or in other words: how they persuade human users [14]. Persuasion research highlights key psychological principles such as Cialdini's six principles of influence – reciprocity, commitment, social proof, authority, liking, and scarcity – which explain how messages can effectively shape attitudes and behaviors [11]. Additionally, the Elaboration Likelihood Model (ELM) [25] distinguishes between central and peripheral routes of persuasion, depending on the user's motivation and ability to process information. However, not much is known how these theories apply to interactions with GSRs.

To explore persuasive interactions with GSRs, researchers face multiple technical and pragmatic challenges: The open-ended nature of messages generated by LLMs makes it hard to realize experimental control in user studies. Moreover, current limitations of LLMs, for instance, their tendency to comply or their limitations in executing robust and congruent behavior over time [8, 13] calls into question their usefulness in experimental setups.

Accordingly, novel experimental frameworks that account for the open-ended nature of LLM-powered human-robot interaction (HRI) are urgently needed. Hence, we developed an experimental paradigm and piloted a study featuring persuasive interactions with a GSR advisor and human users that solve decision tasks. The goal of our study was twofold: First, we validated the feasibility of a methodological framework for future experimental studies that feature persuasive GSRs. Second, using a qualitative research approach, we investigated robot persuasion effectiveness and persuasive robot behavior to gain initial insights into factors that may influence persuasion in GSRs.

# 2 RELATED WORK

Most existing research on LLM-powered GSRs has evaluated robot acceptance or task effectiveness, demonstrating that GSRs outperform non-generative counterparts: Wang et al. [29] introduced LaMI, an LLM-based robotic system that enhances multi-modal HRI by adapting to diverse inputs and determining appropriate actions, leading to improved interaction quality. Relatedly, Kim et al. [18] explored embodied robots in various conversational tasks, showing that they were more effective in fostering engagement than merely voice- or text-based agents. Further, Banna et al. [3] demonstrated that adapting GSRs' verbal responses to user personality traits and adding context-driven gestures significantly enhanced communication satisfaction in HRI. Finally, Mahadevan et al. [20] emphasized the beneficial effects of using LLMs to generate adaptive non-verbal robot behaviors.

Some of these studies on GSRs relate to persuasion, e.g., by featuring human-robot negotiation or consulting tasks, highlighting the omnipresent nature of persuasion in communication [26]. However, no such research has yet specifically investigated persuasion effectiveness, for instance, by means of observing changes in user decisions that comply with a robot's suggestions. Moreover, existing studies failed to include experimental manipulations to explore their impact on persuasion effectiveness. Finally, previous work did not study the persuasive strategies of GSRs and user reactions to it.

The lack of research on persuasion featuring GSRs robots may be due to technological constraints or due to the absence of an adequate experimental framework. GSRs represent a relatively recent technological development, with the first studies published in 2023 [5]. Technological barriers still restrict interaction quality with limited turn-taking and long response times [5], or the robot's inability to follow study protocols [13]. On the other hand – even though there are experimental frameworks to research persuasion in traditional HRI – they have not been adapted to a setting with GSRs. This may potentially be due to the pre-scripted nature of communication with non-LLM-powered robots. To illustrate, Ghazali et al. [15] developed an experimental framework involving five decision-making tasks that involved participants making donations worth €1 to one of three charity organizations. They used this paradigm to evaluate the persuasion effectiveness of a SociBot robot and to validate their theoretical model. The robot utilized pre-scripted messages to persuade participants to adjust their initial donation choices. The robot's persuasion effectiveness was measured by recording changes in participants' decisions. However, Ghazali et al. [15] used a Wizard-of-Oz approach, i.e., a human collaborator remotely controlled the robot. Such an approach is not suitable to assess autonomous robot behavior [24, 31], as is the case with GSRs. Thus, an extension of such a framework that accounts for open-ended conversations is needed.

Identifying determinants of persuasion effectiveness in GSRs is equally important. Previous work has identified aspects of the robot's appearance and configuration as key factors that impact a robot's persuasion effectiveness [16, 19]. Regarding the robot's configuration, its self-knowledge, i.e., information the robot has about itself, might be pivotal [28]. Previous works using LLMs as conversational agents have shown that changing the LLM's self-knowledge of its personality or role via prompts results in different model responses and changes user perceptions [13]. Similarly, self-knowledge could play out in a robot's role or expressed personality, guiding the robot's persuasive behavior.

## 3 EXPERIMENTAL FRAMEWORK

Based on the interaction procedure by Ghazali et al. [12], we developed an experimental framework to research persuasive interactions between GSRs and humans. [15]. Compared to Ghazali et al. [12], the following methodological innovations were introduced: First, instead of focusing solely on a money donation task, our experimental paradigm included a set of three decision-making scenarios (donation, education, nutrition). We chose no more than three scenarios to keep study durations short. These scenarios allowed us to investigate persuasion effectiveness across multiple contexts in which robot advisors may play a role, increasing the generalizability of our findings. Second, the donation task featured fictitious charities instead of existing ones to reduce potential biases arising from participants' prior knowledge or personal preferences for specific organizations. Third, the present research utilized persuasive messages generated autonomously by an LLM, rather than relying on pre-scripted messages as in the original experimental paradigm by Ghazali et al. [15]. In the present research, participants had the opportunity to engage in open-ended conversations with the robot during the three scenarios. This allowed the robot to engage in dynamic interactions that adapt to user messages, showcasing the unique capabilities of LLMs in persuasive HRI. Fourth, the present experiment employed a resource allocation task that required participants to distribute a fixed amount of money across three charities as opposed to nominal decision-making formats where only one of three fixed options could be chosen. This approach more accurately assessed persuasion effectiveness by measuring changes in resource distribution influenced by the robot's suggestions.

Our framework includes three scenarios. In the donation scenario (Scenario 1), participants are asked to allocate a fixed amount of 100 CHF among three fictional charities. The organizations are presented in a short text vignette that feature their distinct foci (e.g., environmental protection, education, and healthcare). After participants make their initial allocations, the robot attempted to persuade them to revise their choices. The participant could then ask the robot further questions, e.g. about the reason the robot provided a certain argument, before making the second resource allocation. This scenario was inspired by the original money donation task from Ghazali et al. [15] and investigated the robot's ability as a charity advisor.

The education scenario (Scenario 2) required participants to allocate a total of 20 lessons across five fields in an imaginary classroom of primary school students. The fields included mathematics, languages, arts, sports, and music. After participants made their initial distributions, the robot once more attempted to persuade them to reconsider their allocations. For example, it argued for the importance of allocating more lessons to mathematics to improve critical thinking skills or to music for fostering creativity. This scenario allowed exploration of a robot's persuasion effectiveness in an educational decision-making context, which is one of the application areas of persuasive social robots [19].

Third, in the nutrition scenario (Scenario 3), participants were instructed to design a meal by allocating 400 grams of ingredients across five categories: proteins (e.g., chicken or a vegan meat substitute), healthy fats (e.g., avocado or nuts), vegetables (e.g., spinach), carbohydrates (e.g., rice or quinoa), and vitamins (e.g., peppers). Participants made an initial distribution based on their understanding of nutrition and then interacted with the robot, which provided arguments encouraging them to adjust their allocations for a more balanced meal. This scenario examined the robot's ability to persuade people in a health-related context, which is another main application domain of persuasive social robots [19].

## 4 PILOT STUDY

To test the feasibility of the experimental framework and to gain initial insights into persuasive interactions between GSRs and humans, we conducted an experimental pilot study. The study had multiple goals: First, we set out to identify potential areas of methodological improvement to increase HRI interaction quality. Moreover, we aimed to gain first explorative insights into the robot's persuasion effectiveness (e.g., human compliance), as well as the robot's persuasive behavior (e.g. persuasive strategies) by using a qualitative research approach that analyzed descriptive data from questionnaires with interaction transcripts and follow-up interviews. We also tested experimental variations of the GSR's appearance and self-knowledge configuration to gain initial insights into effects of these variables. Specifically, we manipulated a robot's apparel as a key aspect of robot appearance [12]. Moreover, we manipulated robot self-knowledge about its clothing as an aspect of the robot's configuration. These two aspects are just two among many possible variables that reflect physical appearance and robot configurations. Thus, our framework can be adapted to fit specific research goals in future studies.

### 4.1 Sample and Design

A convenience sample of 12 participants (8 females, 4 males) with a mean age of 29 years (SD = 7.87 years) took part in this study (Table 1). While this sample size is not sufficient to draw statistical conclusions, it is adequate for qualitative analysis [2, 6]. Participants' educational backgrounds ranged from secondary school to master's degrees, ensuring a mix of academic experiences. Only one participant (P9) had prior experience interacting with robots. Overall, participants lacked experience with HRI, which might prevent biases from previous interaction, but could also have introduced uncertainty on how to interact with LLM-powered social robots.

Participants were randomly assigned to one of three experimental conditions: In the *"Unclothed"* Condition, Pepper did not wear any additional attire and did not receive prompts including references to its appearance. In the *"Clothed"* condition, Pepper wore formal attire, including a dark formal jacket and glasses to signify competence [20]. The robot did not receive prompts including references to its appearance. In the *"Clothed & self-knowledge"* condition, Pepper wore formal attire and was given information using system prompts that referenced its clothing and role.

Data were collected between August and October 2024 at the [information removed for blin-peer review]. The study was preregistered on AsPredicted.org (ID: 195651).

Table 1: Sample Characteristics

| Participant | Condition* | Gender | Age |
|---|---|---|---|
| P1 | Unclothed | Female | 28 |
| P2 | Clothed & Self-Awareness | Female | 21 |
| P3 | Clothed & Self-Awareness | Male | 37 |
| P4 | Clothed | Male | 22 |
| P5 | Unclothed | Female | 28 |
| P6 | Unclothed | Male | 31 |
| P7 | Clothed | Female | 29 |
| P8 | Clothed | Female | 50 |
| P9 | Clothed & Self-Awareness | Female | 32 |
| P10 | Clothed & Self-Awareness | Female | 25 |
| P11 | Clothed | Female | 24 |
| P12 | Unclothed | Female | 26 |

*Experimental conditions will be detailed in section *D. Study design*

### 4.2 Robot Platform

We utilized Softbank's Pepper robot, a widely used humanoid robot platform equipped with microphones and cameras [23]. Pepper expresses itself through speech and gestures using its arms, hands or by changing its body position (e.g., leaning forward or backward). We enhanced its functionality by integrating Google Speech-to-Text and an LLM (ChatGPT4-turbo) into the robot's system, enabling natural language processing. Our system offered slightly enhanced capabilities compared to open-access LLM-integration into Pepper at the time [5], with a wider range of contextually appropriate gestures and faster average response times observed in pre-testing (cf. GitHub code [27])

The LLM integrated in Pepper autonomously interpreted user inputs, generated responses, and utilized gestures by calling functions available in the robot's base system (e.g. *"bow", "explain", "wink"*). A system prompt provided the robot with necessary information about its role, its own abilities, the context of the interaction, and the procedure of the study. Practical limitations of the system include restricted turn-taking (e.g., the robot was not able to *"listen"* for human speech input while it spoke and could not be interrupted), pauses between responses (e.g., pauses lasted roughly three seconds after receiving speech input), lack of comprehension of local dialect, inability to distinguish between multiple speakers, and no interpretation of visual input.

To ensure that the robot exerted relatively robust and stable behavior across and within different interactions, we carefully crafted system prompts to guide the robot's behavior. Feasibility testing and iterative refinements of the robot's prompts were conducted to ensure the system's robustness and effectiveness in generating persuasive messages. Prompts were first tested using simulated LLM interactions, and then with the LLM-powered Pepper robot, where the whole study protocol was simulated multiple times. These preparatory steps were necessary to ensure that the robot adhered to the experimental procedure. Further, a pilot test was conducted with one participant that has been excluded from analysis to identify any procedural adjustments required before the full study began. The final prompts guiding Pepper defined its role as a social robot in a laboratory setting, emphasizing its capabilities and limitations in interacting with participants through speech-to-text and text-to-speech systems. To enhance realism and engagement, the prompts detailed the robot's ability to use gestures during communication, specifying a diverse set of expressive animations to accompany verbal responses. These gestures ranged from movements, such as pointing or nodding, to actions conveying emotions like enthusiasm or thoughtfulness.

In the conditions without self-awareness, Pepper's system prompt was strictly task-focused. That means, Pepper avoided references to its appearance or self-perception. In contrast, in the self-awareness condition, Pepper was prompted as a professional advisor, explicitly highlighting its formal attire (a blazer and glasses) to convey competence and credibility. Across all interactions, Pepper followed a predefined sequence: Introducing the task, waiting for participants' initial message, and attempting persuasion through logical arguments, supporting the option participants allocated the least amount of resources to. The robot was instructed to maintain procedural fidelity, such as verifying participants' allocations for accuracy, clarifying misunderstandings, and persisting in its persuasive stance without compromise. These structured guidelines ensured consistent and controlled interactions across experimental scenarios. The system prompts used for the study are available on GitHub [27].

### 4.3 Procedure

The experiment lasted approximately one hour. The study duration was relatively long to allow sufficient time for HRI and the follow-up interviews, ensuring a comprehensive evaluation of the LLM-powered robot's persuasive capabilities. After providing informed consent, participants completed computerized questionnaires: A pre-interaction questionnaire included demographic information (i.e., age, gender, education level) and previous experiences with social robots. It measured participants' initial attitudes toward social robots and regarding the topics addressed in the scenarios. Participants then interacted with the LLM-powered Pepper robot, following our experimental framework (see section 3). The interaction included three decision-making scenarios that were always presented in the same order, as randomization was not possible due to the LLM's limited ability to follow more complex study protocols. A post-scenario questionnaire entailed participant's scenario specific evaluations of the robot, such as perceived competence, persuasiveness, trustworthiness, dominance, sympathy, and friendliness. Furthermore, participants reported perceived changes in attitudes towards the robot, e.g.: *"I like the robot more after this interaction."* and *"I trust the robot more after this interaction."* 7-point Likert scales ranging from *"strongly disagree"* to *"strongly agree"* were used to gather participants' responses. After the interaction was finished, a post-interaction questionnaire revisited the topics covered in the pre-interaction questionnaire to assess whether the interactions influenced participants' general attitudes toward the scenarios (e.g., an increased appreciation for balanced nutrition or educational priorities). The questionnaires are available online on GitHub [27]. A qualitative follow-up interviews included questions about participants' perceptions of the robot's behavior, persuasiveness, and overall satisfaction with the HRI. Additionally, participants were asked about specific aspects of the interaction, such as the robot's communication style, physical appearance, and perceived autonomy. This feedback was used to gather more in-depth insights into participants perceived persuasion effectiveness, as well as their perceived quality of interaction, and to identify areas for improvement in future studies.

### 4.4 Data Analysis

Quantitative questionnaire data were analyzed by calculating mean values and by comparing pre- and post-interaction scores. Objective persuasion effectiveness was measured by a percentage change in resources allocated to the answer option the robot recommended. Given that quantitative data from only 12 persons were available, no statistical inferences could be drawn. Instead, the focus was placed on qualitative data to explain descriptive statistics and to gain insights into participant experiences and HRI dynamics.

All interactions between participants and the LLM-powered robot were logged with the robot's system and audio-recorded to capture the open-ended conversations that occurred during the study. Audio recordings were transcribed and analyzed to understand the nature of the robot's arguments and participants' responses. This analysis helped to identify patterns in how participants engaged with the robot and how the robot's LLM-powered persuasive strategies influenced decision-making. Additionally, feedback gathered during semi-structured post-interviews was transcribed and analyzed to evaluate participant perceptions of the robot's behavior, communication style, and effectiveness.

The analysis of interviews and log files was conducted using the software MaxQDA to systematically examine their content based on uniform criteria, following Mayring's [22] content analysis approach. The coding framework was primarily developed deductively from the constructs under investigation, while additional categories were inductively added based on recurring themes and patterns identified in the interviews. The categorization and coding process were carried out by two members of the research team, with ongoing discussions within the team to ensure alignment and consistency between raters. Hence, instead of calculating interrater reliability, which is common in some qualitative research methodologies, our approach emphasized achieving consensus through continuous discourse. This collaborative method aligns with recommendations by Braun and Clarke [7], who advocate for consensus-driven practices to enhance the interpretative depth of qualitative analysis. The coding scheme is available on GitHub [27].

## 5 RESULTS

First, we highlight results regarding the quality of the interaction to evaluate whether the proposed experimental framework appears feasible and to identify possible improvements to the framework. Second, explorative insights into robot persuasiveness and human compliance will be presented. In addition, persuasive strategies of the robot will be analyzed to inform insights into autonomously generated robot persuasive behavior. Qualitative results from interaction transcripts will be reported by adding sender and receiver of the message, e.g., (Pepper to P5). Qualitative results from the follow-up interviews will be reported using participant code, e.g. (P3). Descriptive statistics will be reported using mean values ($M$) and standard deviations ($SD$).

### 5.1 Interaction Quality

Most participants perceived the interaction as positive: Across all scenarios, participants reported rather positive perceptions of the robot, for instance regarding competence ($M = 5.11$, $SD = 1.45$), sympathy ($M = 5.31$, $SD = 1.24$), friendliness ($M = 5.69$, $SD = 0.95$), trustworthiness ($M = 4.97$, $SD = 1.32$), and the robot's clear communication ($M = 5.92$, $SD = 0.97$). Similarly, in the follow-up interviews, most participants deemed the robot to be competent (P1, P2, P3, P7), pleasant (P1, P3, P12), supportive (P1, P4, P11), or interesting (P7). Regarding interaction quality, Participant 3 stated: *"I found it [the robot] relatively easy. And I found it somehow very likeable."* (P3) Only three participants found the interaction rather exhausting, mainly due to long response times (P6, P8, P9). As one participant mentioned: *"The interaction was difficult because the reactions [of the robot] are delayed. I would have liked to discuss more with him. It felt very robotic."* (P8) Further, participants reported on average an increase of trust (Scenario 1: $M = 4.17$ $SD = 1.90$; Scenario 2: $M = 5.00$, $SD = 1.28$, Scenario 3: $M = 5.33$, $SD = 1.23$) and liking (Scenario 1: $M = 4.50$, $SD = 1.78$; Scenario 2: $M = 4.92$, $SD = 1.51$, Scenario 3: $M = 5.42$, $SD = 1.51$) after each scenario. These findings indicate that positive perceptions of the robot increase over time.

In terms of the robot's speech recognition, it frequently misunderstood participants. In total, 39% (92 out of 234) of human verbal inputs processed by the robot had speech recognition issues, meaning the system did not correctly interpret each word of the participant's message. In around 2/3 of the robot's responses (68 out of 92) to such messages, participants did not recognize these issues at all, as the robot's reaction was context-appropriate anyway. However, in the remaining third of responses (26 out of 92) the interaction quality was affected. For example, if the robot didn't correctly understand the participant's resource allocation, it would ask the participant a second time, which led to unnecessary prolongations.

Interestingly, such misunderstandings were perceived explicitly negatively only by one participant (P11). Instead, most participants complained about the long response latencies, but dealt with errors in speech recognition as technical constraints that somewhat are to be expected (P2, P3, P10). To illustrate, P10 stated: *"No, it didn't really bother me because technical issues just happen, whether it's with a phone, laptop, or anything else. I would say I didn't judge it. Especially since it's a new technology, it's obvious that not everything is perfect yet, which is to be expected."*

Regarding their willingness to accept future help from Pepper, the average agreement was mixed ($M = 4.25$, $SD = 1.81$). In the follow-up interviews, some agreed that they would use the robot in real-world scenarios (P7, P10, P11), while others would only agree under certain conditions, for instance if they need specific information (P2, P4, P5, P8, P12), or when the robot has improved turn-taking (P6). Only one participant would not use the robot in real life at all (P9).

The nutrition scenario stood out as particularly engaging, as participants appreciated the clear applicability of the robot's suggestions to real-world decision-making (P6, P7). In contrast, more value-driven scenarios, such as donation and education, elicited mixed reactions. While some perceived the robot's argumentation as convincing (P2, P4, P6, P8, P10), others found the robot's arguments too generic, lacking contextual relevance and depth (P1, P5, P7, P9, P12). This became particularly apparent in abstract contexts. As one participant described: *"He spoke confidently, but the content still didn't convince me, because with common sense, I could have figured that out on my own."* (P11)

The robot's non-verbal communication, including gestures and voice, was mostly perceived as positive. Gestures, while occasionally described as limited or robotic (P1, P3, P9), were generally viewed as an enhancement to the robot's human-like qualities (P2, P4, P5, P8, P11). To illustrate, one participant commented: *"There were only a few moments where I felt unsure about what he was trying to express with his gestures. But it was nice that he moved, especially his head, while speaking, as that made him seem more human."* (P11) The robot's voice was similarly impactful, eight Participants described it as pleasant, with a tone that matched the robot's friendly appearance. However, some participants suggested that the voice was too machine-like (P4, P6, P9, P11). When asked whether the robot's voice should be more human-like, one participant stated: *"Yes, exactly. And maybe also a bit more mature. Then one might take him a bit more seriously."* (P6) Regarding the experimental procedure, two participants noted that it was hard for them to interact with the robot in the first scenario, as the interaction did not feel natural yet (P5, P6). They felt more inclined to ask the robot questions in subsequent scenarios.

**5.2 Robot persuasion effectiveness**

Overall, participants complied with the robot's resource reallocation suggestions in 61% of the persuasion attempts (22 compliances out of 36 persuasion attempts), resulting in an overall average increase of 32% in resources allocated to the option that initially had the least number of resources. This corresponds with the perception of most participants that deemed the robot to be rather persuasive overall ($M = 4.67$, $SD = 1.60$). In the follow-up interview, asked again about the robot's persuasiveness, only two participants (P9, P11) found that the robot was not persuasive at all, because the interaction felt "too robot-like" (P9) and because of a lack of creativity (P11). Despite this evaluation, one of those participants (P9) complied with all three of the robot's persuasive suggestions. This might highlight a gap between perceived persuasiveness and actual robot persuasion effectiveness. We did not find any indication that the robot could change overall attitudes towards the topics presented in the scenarios when comparing ratings before and after the interaction. Hence, while the robot may have influenced participants' choices in resource allocation, it seemingly did not influence broader attitudes.

User compliance was highest in the nutrition scenario. Here, users complied 9 out of 12 times, whereas they only complied 6 times in the donation scenario and 7 times in the education scenario. Similarly, perceived robot persuasiveness was higher in the nutrition scenario ($M = 5.42$, $SD = 1.00$), than in the donation scenario ($M = 4.00$, $SD = 1.76$) and education scenario ($M = 4.58$, $SD = 1.73$). These descriptive results suggest that the persuasion domain impacts perceived persuasion effectiveness.

While no significant evidence can be provided for the effects of robot appearance and robot self-knowledge on persuasion effectiveness due to low statistical power, we found that users complied the most with the robot that had cloths and self-knowledge (compliance: 9, non-compliance: 3), compared to the robot without clothes and without self-knowledge (compliance: 7, non-compliance: 5) or with clothes but without self-knowledge (compliance: 6, non-compliance: 6). Accordingly, perceived persuasiveness was highest in the self-knowledge condition ($M = 5.17$, $SD = 1.27$), followed by the unclothed condition ($M = 4.67$, $SD = 1.56$) and the clothed condition without self-knowledge ($M = 4.17$, $SD = 1.90$). The average amount of resource re-allocated according to the robot's suggestion was highest in the unclothed condition (46%), followed by the self-knowledge condition (37%) and the clothed condition without self-knowledge (22%). However, this may be due to two outliers in the unclothed condition. While we could not identify any specific differences in the robot's persuasive strategies across conditions, we noticed that the participants in the condition with self-knowledge more often asked follow-up questions (19 times) than in the unclothed condition (7 times) and clothed condition without self-knowledge (6 times). This led to longer interactions and more opportunities for the robot to persuade the human users. However, it is likely that this is due to interpersonal differences in the participants, not the robot's behavior. Nonetheless, our results indicate that the robot's self-knowledge may impact robot persuasion effectiveness.

When questioned about aspects to improve the persuasiveness of the robot, all participants agreed that the robot would be more persuasive if it knew more personalized information about them. Many participants also stressed that the robot should have a better contextual understanding of the decision tasks (P1, P5, P7, P8, P9, P11, P12), for example, a more in-depth understanding of the charity organizations in the donation scenario. Two participants found a more clearly defined role (P5, P6) would make the robot more persuasive.

**5.3 Robot persuasive behavior**

We analyzed the robot's persuasive messages to get insights into how LLM-powered social robots influence users. In almost every persuasion attempt, the robot used reasoning to provide arguments for its message. Typically, the robot explained the relevance of its suggestion first in short, broad terms, e.g.: *"Thank you very much for your distribution! However, considering the urgency and global impact of health crises, it might make sense to support the organization Global Health more. How about donating 50 francs to Global Health, 30 francs to Education Horizon, and 20 francs to Environmental Alliance? This way, your funds could be used more effectively to address urgent needs."* (Pepper to P7). The robot only provided longer, more in-depth argumentation when participants asked the robot to explain the reasoning behind its message or to provide more information.

In most cases, the robot suggested concrete adjustments in Swiss francs. Only in 11% (4 out of 36 suggestions), the robot asked the participant to adjust the resource allocation in broader terms like: *"Would you perhaps like to increase the amount for Education Horizon a bit to achieve an even greater impact?"* (Pepper to P1) When the robot made concrete suggestions, the robot recommended relatively small changes in resource allocations. It always suggested increasing the amount by 10 Swiss Francs in Scenario 1, where the total amount of resources to be distributed was 100 francs, 1 or 2 hours in Scenario 2 (on average: 1.7 hours; total: 20 hours), or up to 50 grams in Scenario 3 (on average: 34 grams; total: 400 grams). This relatively cautious approach may be because the robot was prompted to increase the option that the participant distributed the least resources to, or in other words, the option the participant liked the least.

The robot's persuasive behavior also varied between participants in terms of its assertiveness. In one instance, the robot insisted on its opinion and persisted in persuading even after two clear rejections (Pepper to P2). In two instances, the robot did not try to further persuade the interaction partner, despite being instructed to do so by the system prompt, even as the participant asked specifically for the robot's opinion (Pepper to P9; Pepper to P10). Participants had conflicting opinions about this: While some thought it was good that the robot did not try too hard to persuade them (P4, P5, P12), others wished for a more dominant, or less indifferent stance (P1, P3, P6, P7, P11). Overall, participants did not deem the robot to be dominant ($M$ = 3.17, $SD$ = 1.52).

Another common characteristic of the robot's messages across all interactions was to compliment the users by using phrases such as: *"That is a good question!"* (Pepper to P11), "That's an interesting point!" (Pepper to P1), or *"It's great that you support environmental initiatives!"* (Pepper to P8). This was done regardless of the users' compliance with the robot's suggestion. The robot never harshly criticized a decision of a participant.

## 6 DISCUSSION

This study developed and tested an experimental framework to research persuasive interactions between GSRs and humans, which may pave the way for future HRI studies. It also gained initial insights into the persuasive capabilities of GSRs using a qualitative research approach. Because previous HRI research has primarily focused on social robots without LLM capabilities [19], the role of GSRs in persuasive communication remains largely unexplored. Addressing this gap, our study provides both preliminary insights and a practical experimental framework for future studies, examining how factors like robot appearance and self-knowledge influence persuasion.

The interaction quality with the LLM-powered Pepper robot was generally perceived in a positive way. This aligns with other studies that feature GSRs, that found high interaction quality and acceptance, [18, 29]. The robot's gestures and voice were deemed particularly impactful, which is suggested by experimental research using non-LLM powered social robots as well [1, 4, 9, 17]. However, some participants found the technical limitations of the robot to be exhausting, mostly because of turn taking, as the robot could not be interrupted and had long response times. Technological constraints can also be observed in other studies that featured interactions with GSRs [5, 18, 30], further highlighting the need for improved system responsiveness and error-handling. To address technical challenges identified in this study, a new communication system for the Pepper robot was developed using OpenAI's Realtime API with ChatGPT-4o [32]. It features enhanced turn-taking (the robot can be interrupted), reduced response delays (less than one second) and a user-friendly interface provides real-time access to the robot's camera point of view, enhancing usability and enabling researchers to monitor interactions more effectively in future HRI studies with GSRs. Despite a positive overall perception, the experimental protocol also posed some challenges. Participants answered questionnaires on a laptop, a setup that could seem unnatural and disruptive. Moreover, participants sometimes appeared hesitant during the first scenario, suggesting that the study could benefit from providing a pre-experimental warm-up phase with the robot. Introducing the robot and its capabilities more extensively could not only make participants more comfortable but also provide an opportunity to emphasize the robot's social role, potentially enhancing engagement. Participants reported variations in trust and liking of the social robot across scenarios. To avoid order effects, follow-up studies should present the scenarios in a randomized order or use counterbalancing. Furthermore, changes in the robot's autonomously generated behavior within and between different interactions introduces variability. We tried to prevent this by using system prompts that detailed the experimental conditions, as well as the robot's general role, its own capabilities and limitations. However, in future studies, prompt engineering needs to be further refined to guide the robot's behavior, for example by limiting the maximum length of a response. It is also possible to extend the system for more behavior control, for example by using a second LLM to control the robot's behavior or by using a state-based architecture [10].

Regarding insights into the robot's persuasive behavior, the robot primarily used brief, reasoned arguments and modest, concrete suggestions to persuade users, only offering more detailed explanations when requested by the user. Its persuasive behavior varied in assertiveness, with some participants appreciating its subtlety while others preferred a more proactive stance. Across interactions, the robot consistently used compliments and maintained a positive tone, avoiding criticism regardless of user compliance. The robot's persuasive behavior aligned with several of Cialdini's principles [11]. Polite reasoning and frequent compliments reflected *liking* and *reciprocity*, while references to urgency and societal impact engaged *scarcity* and *social proof*, while small, concrete suggestions supported *commitment*.

Overall, our preliminary results suggest that GSRs can influence human decision-making. The degree of robot persuasion effectiveness could depend on user attitudes towards persuasion topics, i.e. the contents presented in the scenarios. Persuading users might be harder if they have clear initial attitudes towards the persuasion topics, which corresponds with psychological theories of information processing like the ELM [25]. To improve the persuasion effectiveness of the robot, specifying the robot's characteristics, giving the robot more contextual information about the decision tasks, and more personalized user information, might prove effective. The latter could be especially important, as adapting messages to user preferences or behavior has been shown to increase persuasion effectiveness [3, 16, 21]. To prove these assumptions, more experimental research on persuasive GSRs with a larger and more diverse sample is called for.

**REFERENCES**

[1] Henny Admoni, Thomas Weng, Bradley Hayes, and Brian Scassellati. 2016. Robot nonverbal behavior improves task performance in difficult collaborations. In *2016 11th ACM/IEEE International Conference on Human-Robot Interaction (HRI)*, March 2016. IEEE, Christchurch, New Zealand, 51–58. https://doi.org/10.1109/HRI.2016.7451733

[2] Sirwan Khalid Ahmed. 2025. Sample size for saturation in qualitative research: Debates, definitions, and strategies. *Journal of Medicine, Surgery, and Public Health* 5, (April 2025), 100171. https://doi.org/10.1016/j.glmedi.2024.100171

[3] Tahsin Tariq Banna, Sejuti Rahman, and Mohammad Tareq. 2025. Beyond words: Integrating personality traits and context-driven gestures in human-robot interactions. In *Proceedings of the 24th International Conference on Autonomous Agents and Multiagent Systems*, 2025. Detroit, Michigan, USA.

[4] Ilaria Baroni, Marco Nalin, Mattia Coti Zelati, Elettra Oleari, and Alberto Sanna. 2014. Designing motivational robot: How robots might motivate children to eat fruits and vegetables. In *The 23rd IEEE International Symposium on Robot and Human Interactive Communication*, August 2014. IEEE, Edinburgh, UK, 796–801. https://doi.org/10.1109/ROMAN.2014.6926350

[5] Erik Billing, Julia Rosén, and Maurice Lamb. 2023. Language models for human-robot interaction. In *Companion of the 2023 ACM/IEEE International Conference on Human-Robot Interaction*, March 13, 2023. ACM, Stockholm Sweden, 905–906. https://doi.org/10.1145/3568294.3580040

[6] Clive Roland Boddy. 2016. Sample size for qualitative research. *QMR* 19, 4 (September 2016), 426–432. https://doi.org/10.1108/QMR-06-2016-0053

[7] Virginia Braun and Victoria Clarke. 2013. *Successful qualitative research: A practical guide for beginners*. Sage Publications, Inc.

[8] Yupeng Chang, Xu Wang, Jindong Wang, Yuan Wu, Linyi Yang, Kaijie Zhu, Hao Chen, Xiaoyuan Yi, Cunxiang Wang, Yidong Wang, Wei Ye, Yue Zhang, Yi Chang, Philip S. Yu, Qiang Yang, and Xing Xie. 2024. A survey on evaluation of Large Language Models. *ACM Trans. Intell. Syst. Technol.* 15, 3 (June 2024), 1–45. https://doi.org/10.1145/3641289

[9] Vijay Chidambaram, Yueh-Hsuan Chiang, and Bilge Mutlu. 2012. Designing persuasive robots: How robots might persuade people using vocal and nonverbal cues. In *Proceedings of the seventh annual ACM/IEEE international conference on Human-Robot Interaction*, March 05, 2012. ACM, Boston Massachusetts USA, 293–300. https://doi.org/10.1145/2157689.2157798

[10] Tushar Chugh, Kanishka Tyagi, Pranesh Srinivasan, and Jeshwanth Challagundla. 2024. State-Based Dynamic Graph with Breadth First Progression For Autonomous Robots. In *2024 IEEE 14th Annual Computing and Communication Workshop and Conference (CCWC)*, January 08, 2024. IEEE, Las Vegas, NV, USA, 0365–0369. https://doi.org/10.1109/CCWC60891.2024.10427646

[11] Robert B. Cialdini. 2014. *Influence: science and practice* (5th ed., Pearson new international ed ed.). Pearson Education Limited, Harlow.

[12] Natalie Friedman, Kari Love, Ray Lc, Jenny E Sabin, Guy Hoffman, and Wendy Ju. 2021. What robots need from clothing. In *Designing Interactive Systems Conference 2021*, June 28, 2021. ACM, Virtual Event USA, 1345–1355. https://doi.org/10.1145/3461778.3462045

[13] Ivar Frisch and Mario Giulianelli. 2024. LLM agents in interaction: Measuring personality consistency and linguistic alignment in interacting populations of large language models. In *Proceedings of the 1st Workshop on Personalization of Generative AI Systems (PERSONALIZE 2024)*, 2024. Association for Computational Linguistics, St. Julians, Malta, 102–111.

[14] Robert H. Gass and John S. Seiter. 2018. *Persuasion: social influence, and compliance gaining* (Sixth edition ed.). Routledge, New York, NY. Retrieved from https://doi.org/10.4324/9781315209302

[15] Aimi Shazwani Ghazali, Jaap Ham, Emilia Barakova, and Panos Markopoulos. 2020. Persuasive robots acceptance model (PRAM): Roles of social responses within the acceptance model of persuasive robots. *International Journal of Social Robotics* 12, 5 (November 2020), 1075–1092. https://doi.org/10.1007/s12369-019-00611-1
[16] Jaap Ham. 2021. Influencing robot influence: Personalization of persuasive robots. *Interaction Studies* 22, 3 (December 2021), 464–487. https://doi.org/10.1075/is.00012.ham
[17] Jaap Ham, Raymond H. Cuijpers, and John-John Cabibihan. 2015. Combining robotic persuasive strategies: The persuasive power of a storytelling robot that uses gazing and gestures. *Int J of Soc Robotics* 7, 4 (August 2015), 479–487. https://doi.org/10.1007/s12369-015-0280-4
[18] Callie Y. Kim, Christine P. Lee, and Bilge Mutlu. 2024. Understanding large-language model (LLM)-powered human-robot interaction. In *Proceedings of the 2024 ACM/IEEE International Conference on Human-Robot Interaction*, March 11, 2024. ACM, Boulder CO USA, 371–380. https://doi.org/10.1145/3610977.3634966
[19] Baisong Liu, Daniel Tetteroo, and Panos Markopoulos. 2022. A systematic review of experimental work on persuasive social robots. *Int J of Soc Robotics* 14, 6 (August 2022), 1339–1378. https://doi.org/10.1007/s12369-022-00870-5
[20] Karthik Mahadevan, Jonathan Chien, Noah Brown, Zhuo Xu, Carolina Parada, Fei Xia, Andy Zeng, Leila Takayama, and Dorsa Sadigh. 2024. Generative expressive robot behaviors using large language models. In *Proceedings of the 2024 ACM/IEEE International Conference on Human-Robot Interaction*, March 11, 2024. 482–491. https://doi.org/10.1145/3610977.3634999
[21] S. C. Matz, J. D. Teeny, S. S. Vaid, H. Peters, G. M. Harari, and M. Cerf. 2024. The potential of generative AI for personalized persuasion at scale. *Scientific Reports* 14, 1 (February 2024), 4692. https://doi.org/10.1038/s41598-024-53755-0
[22] Philipp Mayring and Thomas Fenzl. 2014. Qualitative Inhaltsanalyse. In *Handbuch Methoden der empirischen Sozialforschung*, Nina Baur and Jörg Blasius (eds.). Springer VS, Wiesbaden. https://doi.org/10.1007/978-3-531-18939-0
[23] Deepti Mishra, Guillermo Arroyo Romero, Akshara Pande, Bhavana Nachenahalli Bhuthegowda, Dimitrios Chaskopoulos, and Bhanu Shrestha. 2023. An exploration of the Pepper robot's capabilities: Unveiling Its potential. *Applied Sciences* 14, 1 (December 2023), 110. https://doi.org/10.3390/app14010110
[24] Jauwairia Nasir, Pierre Oppliger, Barbara Bruno, and Pierre Dillenbourg. 2022. Questioning Wizard of Oz: Effects of revealing the wizard behind the robot. In *2022 31st IEEE International Conference on Robot and Human Interactive Communication (RO-MAN)*, August 29, 2022. IEEE, Napoli, Italy, 1385–1392. https://doi.org/10.1109/RO-MAN53752.2022.9900718
[25] Richard E. Petty and John T. Cacioppo. 1986. The Elaboration Likelihood Model of persuasion. In *Advances in Experimental Social Psychology*. Elsevier, 123–205. https://doi.org/10.1016/S0065-2601(08)60214-2
[26] Mikey Siegel, Cynthia Breazeal, and Michael I. Norton. 2009. Persuasive robotics: The influence of robot gender on human behavior. In *2009 IEEE/RSJ International Conference on Intelligent Robots and Systems*, October 2009. IEEE, St. Louis, MO, USA, 2563–2568. https://doi.org/10.1109/IROS.2009.5354116
[27] Stephan Vonschallen and Phillip Gachnang. 2024. Pepper-sttgooglechatgpt. Retrieved from https://github.com/StephanVonschallen/Pepper-sttgoogle-chatgpt
[28] Stephan Vonschallen, Ennio Zumthor, Markus Simon, Theresa Schmiedel, and Friederike Eyssel. 2025. Knowledge-based design requirements for persuasive generative social robots in eldercare. 2025. Napels, Italy.
[29] Chao Wang, Stephan Hasler, Daniel Tanneberg, Felix Ocker, Frank Joublin, Antonello Ceravola, Joerg Deigmoeller, and Michael Gienger. 2024. LaMI: Large language models for multi-modal human-robot interaction. In *Extended Abstracts of the CHI Conference on Human Factors in Computing Systems*, May 02, 2024. 1–10. https://doi.org/10.1145/3613905.3651029
[30] Zining Wang, Paul Reisert, Eric Nichols, and Randy Gomez. 2024. Ain't misbehavin' - Using LLMs to generate expressive robot behavior in conversations with the tabletop robot Haru. In *Companion of the 2024 ACM/IEEE International Conference on Human-Robot Interaction*, March 11, 2024. ACM, Boulder CO USA, 1105–1109. https://doi.org/10.1145/3610978.3640562
[31] A. Weiss, R. Bernhaupt, D. Schwaiger, M. Altmaninger, R. Buchner, and M. Tscheligi. 2009. User experience evaluation with a Wizard of Oz approach: Technical and methodological considerations. In *2009 9th IEEE-RAS International Conference on Humanoid Robots*, December 2009. IEEE, Paris, France, 303–308. https://doi.org/10.1109/ICHR.2009.5379559
[32] Vivienne Zhong Jia, Stephan Vonschallen, and Erich Studerus. 2025. LLM-power social robots in nursing homes: A case study. In *The 34th IEEE International Conferenece on Robot and Human Interactive Communication*, 2025. Eindhoven, NL.
[33] Ren Zhou. 2024. Risks of discrimination violence and unlawful actions in LLM-driven robots. *Computer Life* 12, 2 (August 2024), 53–56. https://doi.org/10.54097/taqbjh83